\documentclass[11pt]{article}
\usepackage[final]{acl}

\usepackage{times}
\usepackage{latexsym}
\usepackage[T1]{fontenc}
\usepackage[utf8]{inputenc}
\usepackage{microtype}
\usepackage{inconsolata}
\usepackage{tcolorbox}
\tcbuselibrary{breakable}
\usepackage{graphicx}
\usepackage{amsmath}
\usepackage{amssymb}
\usepackage{booktabs}
\usepackage{multirow}
\usepackage{makecell}
\usepackage{enumitem}
\usepackage{url}
\usepackage{natbib}
\usepackage{enumitem}
\usepackage{algorithm}
\usepackage{algorithmic}

\title{Fine-Tuned Multi-Agent Framework for Detecting OCEAN\\
in Life Narratives}

\author{
  Rasiq Hussain$^1$ \quad Darshil Italiya$^1$ \quad Joshua Oltmanns$^2$ \quad Mehak Gupta$^1$ \\
  $^1$Southern Methodist University \quad $^2$Washington University in St. Louis \\
  \texttt{\{rasiqh, ditaliya, mehakg\}@smu.edu} \quad \texttt{j.oltmanns@wustl.edu}
}

\begin{document}
\pagestyle{plain}
\maketitle


\begin{abstract}
Accurately assessing personality from text is challenging because traits are latent, context-dependent, and often subtly expressed across long narratives. Large language models (LLMs) offer new opportunities by processing extensive textual contexts, but pretraining of these models can induce latent “personality-like” biases, making single-model inferences inconsistent. We propose a fine-tuned multi-agent framework for detecting OCEAN personality traits, in which sub-agents are conditioned to adopt high, low, or neutral perspectives for each trait through masked language modeling (MLM) and psychometric supervision. A judge LLM aggregates and compares sub-agent outputs to generate final trait predictions, capturing multiple complementary perspectives while mitigating individual model biases. We evaluate the framework on life narrative dataset through quantitative and qualitative experiments, including baselines, ablations, and inference quality analyses. Our approach offers a scalable and interpretable method for text-based personality inference, highlighting the benefits of multi-agent reasoning grounded in psychometric supervision.
\end{abstract}

\section{Introduction}

Personality shapes behavior, decision-making, and interpersonal interactions, making its accurate assessment critical in domains ranging from psychology to personalized services \citep{Youyou2017, Stajner2020, Alshouha2024}. Extracting personality from text is inherently challenging as it is latent, context-dependent, and expressed through subtle linguistic cues, which makes simple keyword- or pattern-based approaches unreliable \citep{Fang2022, Zhu2024}. Long-form narratives, such as life interviews, further complicate inference due to distributed evidence, interactional context, and implicit trait expression.

Recent advances in large language models (LLMs) offer new opportunities for personality detection. LLMs can process and reason over extensive textual contexts, capturing complex semantic patterns without task-specific training \citep{hu2024augmentation, rao2023chatgpt}. Yet, LLMs themselves may exhibit latent "personality-like" tendencies from pretraining, which can bias predictions and lead to inconsistent assessments \citep{gallegos2024bias, jiang2023personallm}. Relying on a single model is therefore prone to incomplete or skewed inferences.

To address these challenges, we propose a structured multi-agent framework in which LLM sub-agents are conditioned to adopt high, low, or neutral perspectives for each OCEAN trait. Each sub-agent produces facet-level inferences with structured supporting evidence, while a judge LLM aggregates and compares these outputs to generate final trait-level predictions. This approach captures complementary perspectives, mitigates biases, and ensures facet and trait-level interpretability.

We further ground our framework in psychometrically validated instruments such as IPIP-NEO \cite{goldberg1999broad}, which provide facet-level behavioral definitions for the Five Factor (OCEAN) traits. These facets serve as fine-grained supervision targets, enabling sub-agents to evaluate narrative evidence at the facet level to support trait-level classification, improving interpretability at both facet and trait levels.

Our design is motivated by two key considerations. First, conditioning LLMs into trait-oriented roles leverages their reasoning capabilities while reducing pretraining-induced idiosyncrasies. Second, personality is relational and multifaceted; aggregating multiple controlled perspectives produces more accurate predictions than single-agent inference alone. By integrating sub-agent specialization with comparative judgment, the framework captures diverse trait-level signals and aligns outputs with psychometric definitions, enhancing both reliability and interpretability.

The main contributions of this work are:  

\begin{enumerate}[noitemsep, topsep=0pt]
    \item We introduce a multi-agent framework for personality assessment that induces high, low, and neutral trait orientations in LLM sub-agents through MLM fine-tuning, role prompts, and prompts including IPIP-NEO facet keys.  
    \item We design a judge-mediated aggregation mechanism that reconciles conflicting trait signals across sub-agents, enabling more accurate personality classification.
    \item We provide comprehensive evaluation through quantitative baselines, ablation studies, semantic alignment analysis, and qualitative use-case case studies to offer insights into how agents in the multi-agent framework infer personality.
\end{enumerate}
\section{Related Work}

\subsection{Personality Assessment from Text}

Computational personality detection from text has progressed from early feature-engineered models to pretrained encoders such as BERT and RoBERTa, typically applied to essays, social media posts, and short-form narratives \citep{mairesse2007using, devlin2018bert, liu2019roberta, celli2013workshop}. These approaches leverage lexical, syntactic, and semantic cues to predict OCEAN traits, but are often limited to short texts, coarse label schemes, and provide limited interpretability \citep{yang2021psycholinguistic}. 

Recent work has explored large language models (LLMs) for personality assessment on longer and more complex texts using prompting-based methods, demonstrating improved ability to capture richer semantic patterns\cite{piastra2025emergent,rao2023chatgpt,hussain-etal-2026-facet}. However, LLMs are prone to hallucinations and exhibit sensitivity to prompt phrasing \citep{gupta2024self, jiang2024personallm, hu2024augmentation}. These limitations motivate the use of structured psychometric supervision, such as IPIP-NEO items, to provide interpretable and theory-grounded guidance for personality inference.

\subsection{Multi-Agent LLMs for Personality Assessment}

Large language models can also exhibit latent behavioral biases inherited from pretraining, which may influence trait predictions and lead to inconsistent personality inferences when relying on a single model \citep{serapio2023personality, gallegos2024bias, salecha2024social}. Structured prompting can partially mitigate these effects \citep{furniturewala2024thinking}, but single-agent approaches often struggle to capture distributed signals and single-perspective reasoning \citep{du2023improving, liang2023encouraging}.

Multi-agent frameworks address these limitations by enhancing reasoning robustness: multiple LLM agents generate complementary perspectives, which a judge model then aggregates \citep{liang2023encouraging, naik2023diversity, zheng2024judging}. For instance, PADO \citep{yeo2024pado} uses personality-oriented agents induced via role prompts to reduce bias and improve consistency in trait inferences.

Building on this approach, we implement a multi-agent framework in which sub-agents are specialized to HIGH, NEUTRAL, and LOW trait orientations through fine-tuning, role prompts, and IPIP-NEO facet keys. Each sub-agent independently identifies trait signals from its orientation, and aggregating their outputs allows the system to capture subtle, distributed cues that a single agent might miss.

\begin{figure*}[t]
    \centering
    \includegraphics[width=\textwidth]{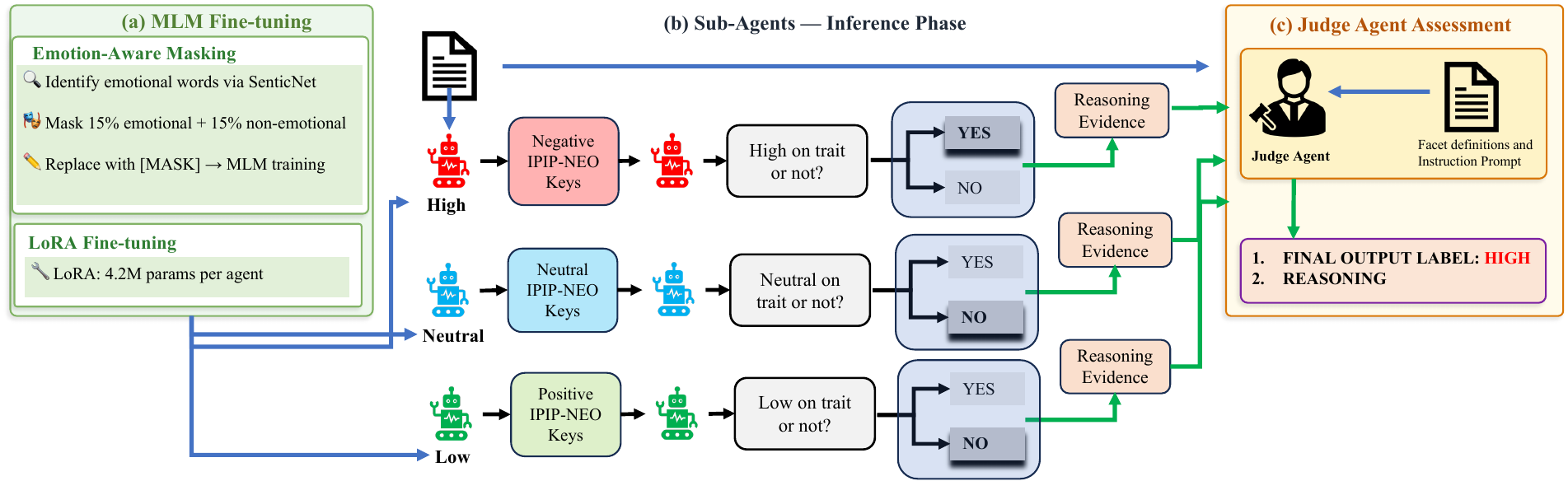}
    \caption{Multi-agent OCEAN Personality Prediction Framework}
    \label{fig:architecture}
\end{figure*}

\section{Methods}
\subsection{Problem Formulation}

Our goal is to predict the Big Five personality traits---Openness, Conscientiousness, Extraversion, Agreeableness, and Neuroticism---from life narratives. Each narrative is annotated with a continuous trait score, which we discretize into three levels: High, Neutral, and Low. Given a narrative $x$, the task is to predict the level of each trait $t \in \{O, C, E, A, N\}$.

To model trait-level variation, we employ a multi-agent framework in which each trait is associated with three sub-agents, one for each trait level: High, Neutral, and Low. This results in a total of 15 sub-agents across the five OCEAN traits. Each sub-agent is fine-tuned to capture distinct linguistic and affective cues indicative of its corresponding trait level and performs a binary classification to determine whether the narrative exhibits the specific trait level it is trained on. A judge agent then aggregates the binary predictions and supporting evidence from all three sub-agents for a given trait to produce a final multi-class decision among High, Neutral, and Low.

Formally, let $y_t \in \{\text{High}, \text{Neutral}, \text{Low}\}$ denote the discrete label for trait $t$. Let $A_t^{(i)}$ denote the $i$-th sub-agent for trait $t$, where $i \in \{\text{High}, \text{Neutral}, \text{Low}\}$. Each sub-agent produces a binary output $b_t^{(i)} \in \{0,1\}$ indicating whether the narrative matches the corresponding trait-level pattern, along with extracted textual evidence $e_t^{(i)}$ and a confidence score $c_t^{(i)}$. The judge agent $J_t$ integrates these signals to predict the final label:
\[
\hat{y}_t = J_t(\{b_t^{(i)}, e_t^{(i)}, c_t^{(i)}\}_{i=1}^3, x).
\]

This hierarchical design allows sub-agents to focus on narrow behavioral cues while the judge agent synthesizes evidence for robust trait inference. An overview of the framework is shown in Figure~\ref{fig:architecture}.

\subsection{MLM Fine-Tuning for Sub-agents}

To induce personality-sensitive representations, we fine-tune all sub-agents using a self-supervised masked language modeling (MLM) objective on life narratives. Rather than relying on personality labels directly, the model learns to predict masked tokens in context, allowing it to internalize emotional and linguistic patterns associated with different trait levels.

For each trait $t$ and trait-level agent $A_t^{(i)}$, we fine-tune a pre-trained LLM on narratives corresponding to the relevant trait level. To capture both emotional and structural linguistic cues, 30\% of tokens in each input sequence are masked: 15\% are emotional words and 15\% are non-emotional words. Emotional words are identified using the SenticNet \cite{cambria2017senticnet} Emotion Lexicon, a large-scale resource mapping words to six emotion dimensions: anger, fear, joy, sadness, disgust, and surprise. The lexicon is filtered to retain only genuinely emotional words, requiring at least one primary emotion with a score $\ge 1.0$ and excluding common non-emotional words, resulting in a set of 2,914 emotion words. Masked tokens are replaced with \texttt{[MASK]}, creating training instances for the MLM objective.

\paragraph{LoRA Fine-tuning:}Each sub-agent is fine-tuned using Low-Rank Adaptation (LoRA) applied to the 
\texttt{q\_proj}, \texttt{k\_proj}, \texttt{v\_proj}, and \texttt{o\_proj} layers 
\cite{liu2024lora}. Instead of updating the full weight matrix, LoRA learns a 
low-rank weight update parameterized by two trainable matrices 
$A \in \mathbb{R}^{r \times d}$ and $B \in \mathbb{R}^{d \times r}$, where 
$r$ is the LoRA rank and $d$ is the hidden dimension. The resulting task vector with task-specific 
weight update is computed as $\tau = \frac{\alpha}{r}BA$, where $\alpha$ is a 
scaling factor. The updated model parameters are calculated as:

\begin{equation}\label{eq:1}
\boldsymbol{\theta}_{\text{merged}} =
\boldsymbol{\theta}_{\text{base}} + \lambda \boldsymbol{\tau},
\end{equation}

where $\lambda$ controls the strength of the task-specific adaptation. Empirically, $\lambda = 0.3$ gave balance between pretrained knowledge and trait adaptation.


\subsection{Sub-agent IPIP-NEO Grounded Prompts}
\label{sec:prompting}
The Five-Factor Model decomposes each trait into six lower-order facets operationalized in the NEO-PI-R \citep{neocosta2008}. Facets capture fine-grained distinct behavioral expressions \citep{deyoung2007between, soto2017bfi2} across each domain. To utilize facet-level information for more precise signals for domain-level trait inference, we incorporate IPIP-NEO facet-level keys. These keys provide positively and negatively correlated IPIP items for each facet \citep{goldberg1999broad}. 

Each sub-agent $A_t^{(i)}$ is grounded in a facet set $\mathcal{F}_t$ corresponding to trait $t$. HIGH-level agents receive positively keyed facet anchors ($\mathcal{F}_t^H$), LOW-level agents receive negatively keyed anchors $\mathcal{F}_t^L$. For NEUTRAL agents, we created moderate cues ($\mathcal{F}_t^N$) by blending the semantic content of positive and negative keys, keeping core behaviors but removing extreme qualifiers or reducing intensity. Refer to Appendix~\ref{app:facets}, Table~\ref{tab:facets_all} for the full list of keys for all agents. This design provides psychometrically grounded supervision and encourages sub-agents to focus on meaningful behavioral signals rather than generic linguistic cues.

Each sub-agent additionally receives a brief role preamble (e.g., “You are an expert in HIGH Neuroticism personality assessment based on the Big Five model”) for further trait orientation. Each sub-agent outputs a structured prediction consisting of a binary decision ($b_t^{(i)}$), supporting evidence ($e_t^{(i)}$), and a confidence score ($c_t^{(i)}$).

\subsection{Judge Agent Assessment}
\label{sec:judge}
For each trait $t$, the judge agent $J_t$ receives (i) the transcript $x$, (ii) structured outputs from the three trait-level sub-agents $\{b_t^{(i)}, e_t^{(i)}, c_t^{(i)}\}_{i=1}^3$, and (iii) trait-specific facet definitions.

\begin{algorithm}[t]
\caption{Multi-agent Personality Inference Framework}
\label{alg:fado}
\begin{algorithmic}[1]
\REQUIRE Narrative transcript $x$
\ENSURE Trait labels $\{\hat{y}_t\}_{t \in \{O,C,E,A,N\}}$

\FOR{each trait $t \in \{O,C,E,A,N\}$}
    \FOR{each level $i \in \{\text{Low, Neutral, High}\}$}
        \STATE Prompt sub-agent $A_t^{(i)}$ with keys $\mathcal{F}_t^{(i)}$
        \STATE Obtain binary decision $b_t^{(i)}$, evidence $e_t^{(i)}$, confidence $c_t^{(i)}$
    \ENDFOR
    \STATE Aggregate sub-agent outputs using judge agent:
    \STATE $\hat{y}_t \leftarrow J_t(\{b_t^{(i)}, e_t^{(i)}, c_t^{(i)}\}_{i=1}^3, x)$
\ENDFOR

\RETURN $\{\hat{y}_t\}_{t \in \{O,C,E,A,N\}}$
\end{algorithmic}
\end{algorithm}

The judge first performs a comparative assessment of the three sub-agents, examining the alignment of each explanation with the transcript, evaluating the coherence and specificity of supporting evidence. It then integrates these assessments to form a holistic evaluation (${y}_t$) of trait expression, along with a brief justification summarizing the key evidence and reasoning. 

Prompt structure for both sub-agent and judge is shared in Supplementary \ref{app:prompt}


\section{Experimental Setup}

\subsection{Data: SPAN Life Narrative Interviews}

Our experiments use life narrative interview transcripts from SPAN \cite{spanstudy_site}. Participants
narrate life experiences through family, work, and formative events in response to open-ended interviewer prompts. We bin the raw ground truth continuous scores into terciles to produce HIGH, NEUTRAL, and LOW labels per trait. We share the label distribution in Table \ref{tab:dataset}. Of 1535 transcripts, we use 50\% transcripts for MLM fine-tuning and 50\% for evaluation. 


\begin{table}[t]
\centering
\small
\caption{Participant distribution across OCEAN terciles. Participants are assigned to HIGH, NEUTRAL, and LOW bins per trait using tercile binning.}
\begin{tabular}{l ccc}
\toprule
\textbf{Trait} & \textbf{HIGH} & \textbf{NEUTRAL} & \textbf{LOW} \\
\midrule
Neuroticism (N)       & 370 & 385 & 356 \\
Extraversion (E)      & 391 & 353 & 367 \\
Openness (O)          & 386 & 375 & 347 \\
Agreeableness (A)     & 377 & 389 & 344 \\
Conscientiousness (C) & 370 & 396 & 343 \\
\bottomrule
\end{tabular}
\label{tab:dataset}
\end{table}


\subsection{Implementation Details}

We use \texttt{Mistral-7B-Instruct} as the backbone for all fifteen sub-agents and \texttt{Qwen} as the judge model. Sub-agents are fine-tuned on a single A100 GPU using Low-Rank Adaptation (LoRA) rank $r=4$, $\alpha=8$, and dropout 0.05, resulting in approximately 4.2M trainable parameters per agent.

During inference, sub-agents are executed sequentially in fp16 precision with GPU memory cleared between runs, while the judge model remains resident to aggregate sub-agent evidence and produce the final prediction. Full implementation details are shared in Supplementary \ref{apx:imp}.

\subsection{Types of Experiments}

We evaluate our framework using a combination of quantitative and qualitative experiments.

\subsubsection{Baseline Models}  
We compare our multi-agent framework against several baselines: (1) a supervised RoBERTa encoder fine-tuned for multi-label OCEAN classification as a strong discriminative baseline, (2) single-agent LLM prompting with chain-of-thought (CoT) using Mistral and Qwen, predicting traits directly without decomposition, and (3) multi-agent setups with, Mistral for both sub-agnets and judge, Qwen for both sub-agnets and judge, and Mistral sub-agents with Qwen judge (our proposed framewrok).

\begin{table}[t]
\centering
\small
\caption{Comparison with baseline models for OCEAN personality prediction (macro-F1).}
\label{tab:baselines}
\setlength{\tabcolsep}{3pt}
\begin{tabular}{l c c c c c }
\toprule
\textbf{Model} & \textbf{O} & \textbf{C} & \textbf{E} & \textbf{A} & \textbf{N}  \\
\midrule
\multicolumn{2}{l}{\textbf{Finetuning}}  & & & &  \\
RoBERTa  & 0.256 & 0.272 & 0.248 & 0.261 & 0.239  \\
\midrule
\multicolumn{2}{l}{\textbf{Single-Agent (CoT)}}& & & &  \\
Mistral  & 0.204 & 0.361 & 0.246 & 0.263 & 0.172  \\
Qwen & 0.381 & \textbf{0.414} & 0.383 & 0.386 & 0.353  \\
\midrule
\textbf{Multi-Agent} & & & & &  \\
All Mistral  & 0.391 & 0.342 & 0.382 & 0.335 & 0.316  \\
All Qwen & 0.403 & 0.331 & 0.347 & 0.357 & 0.347  \\
\makecell[l]{Mistral sub-agents \\+ Qwen Judge} & 
\textbf{0.424} & 0.366 & \textbf{0.416} & \textbf{0.415} & \textbf{0.458}  \\
\bottomrule
\end{tabular}
\end{table}

\subsubsection{Ablation Analysis}  
We perform ablations to evaluate the contribution of each component of the multi-agent framework:

\paragraph{No MLM Fine-Tuning:}  
All fifteen sub-agents use the pretrained Mistral-7B-Instruct-v0.3 model without MLM fine-tuning.

\paragraph{No Sub-Agents:}  
Only the judge agent is used, operating directly on the transcript without input from sub-agents, to assess the importance of sub-agent decomposition.

\paragraph{No Trait-Level Cues:}  
Trait-level specific role prompts are removed, replaced with a generic behavioral analysis prompt, to measure the impact of inference-time role conditioning.

\paragraph{No IPIP-NEO Facet Keys:}  
Facet-level IPIP-NEO facet keys are removed from sub-agent prompts to evaluate the contribution of facet grounding and supervision.

\subsubsection{MLM Evaluation:}  
We evaluate the effectiveness of masked language modeling (MLM) by comparing sub-agent MLM performance before and after fine-tuning using accuracy and perplexity. Test accuracy measures the fraction of masked tokens correctly predicted, while Acc@5 assesses whether the correct token appears among the top 5 predictions. Perplexity quantifies model uncertainty over masked tokens, with lower values indicating better contextual understanding. We will also analyze the personality-specific adaptations after MLM fine-tuning by projecting LoRA-trained personality sub-agent weights into a two-dimensional space.


\subsubsection{Qualitative Analysis}  
To complement quantitative results, we conduct qualitative analysis to examine semantic alignment between sub-agent explanations and their corresponding IPIP-NEO facet keys, evaluating whether sub-agents focus on trait-relevant evidence. We also present a randomly selected case study to illustrate how the judge agent aggregates sub-agent explanations to produce a final decision, and compare this multi-agent reasoning process with a single-agent baseline.
\section{Results and Analysis}

\subsection{Comparison with Baselines}

Table~\ref{tab:baselines} reports macro-F1 performance across OCEAN traits. Our full multi-agent framework achieves the best overall results, improving average macro-F1 by approximately 8\% over the best-performing single-agent model (Qwen with CoT). This highlights the value of aggregating the inference from multiple-agents, rather than relying on a single-agent network. Interestingly, Conscientiousness performs better with a single-agent, suggesting that facet-guided decomposition adds little benefit for this trait due to its cohesive, low-variance facets \citep{deyoung2007between, soto2017bfi2}.


Across alternative multi-agent backbones, the strongest performance is observed when Mistral-7B is used for sub-agents and Qwen-7B is used as the judge. Sub-agents operate on short, facet-specific prompts, where the smaller Mistral model efficiently extracts localized trait evidence. The judge, however, must integrate multiple sub-agent outputs together with the full transcript, producing substantially longer inputs. The larger context capacity of Qwen enables more stable aggregation of this distributed evidence \citep{yuan2024lveval, qwen2025}. In contrast, using Mistral as the judge results in weaker performance when aggregating sub-agent outputs with the full transcript, suggesting that models with smaller context capacity may struggle with long-input aggregation. These results indicate that using different models for evidence extraction and aggregation can be beneficial for complex personality inference tasks.

\begin{table}[t]
\centering
\small
\caption{Per-class accuracy (in \%) of the multi-agent framework.}
\begin{tabular}{l c c c}
\toprule
\textbf{Trait} & \textbf{HIGH}  & \textbf{NEUTRAL} & \textbf{LOW}\\
\midrule
Openness (O)          & 53.7& 27.5& 47.5\\
Conscientiousness (C) & 67.3& 20.6& 25.9\\
Extraversion (E)      &  51.5& 21.5& 51.2\\
Agreeableness (A)     & 60.6& 28.6& 37.6\\
Neuroticism (N)       & 48.8&  41.2& 47.6\\
\bottomrule
\end{tabular}
\label{tab:per_class}
\end{table}

\subsection{Per-Class Performance of the Multi-Agent Framework}

Table~\ref{tab:per_class} reports per-class accuracy for HIGH, LOW, and NEUTRAL labels across OCEAN traits. Accuracy is consistently lowest for NEUTRAL, reflecting the subtlety of moderate trait expressions. In contrast, HIGH and LOW classes show strong performance across traits, indicating that the multi-agent framework effectively captures more distinct trait expressions.

\subsection{Comparison across Ablations}

 Table~\ref{tab:ablation} shows the effect of removing key components from our framework. Removing IPIP-NEO keys results in the largest degradation, with a 15.63\% drop in average F1, highlighting the importance of psychologically grounded lexical anchors.

 Eliminating the MLM component leads to an 11.30\%, while removing trait-level cues from role prompt causes a comparable 10.58\% drop, demonstrating that LoRA-based fine-tuning through MLM task and concise, explicit trait guidance in role prompts both contribute to performance. 
 
 Removing sub-agents (“No Sub-Agents”) results in a 7.93\% drop in overall performance, confirming that agent-level specialization generally improves trait prediction. The full multi-agent configuration achieves the highest macro F1, confirming the advantage of combining specialization decomposition, psychometric grounding, and adaptation.

\begin{table}[t]
\centering
\small
\caption{Ablation results (macro-F1) for the multi-agent framework. Bold indicates the best performance for each trait.}
\setlength{\tabcolsep}{4pt}
\begin{tabular}{ l c c c c c c }
\toprule
\textbf{Ablation} & O & C & E & A & N & \textbf{Avg} \\
\midrule
No MLM &  .389 & .369 & .392 & .342 & .351 & .369 \\
No Trait-level Cues & .408 & .378 & .372 & .356 & .347 & .372 \\
No IPIP-NEO keys & .403 & .328 & .364 & .328 & .331 & .351 \\
No sub-agents & .381 & \textbf{.414} & .383 & .386 & .353 & .383 \\
Full Multi-Agent & \textbf{.424} & .366 & \textbf{.416} & \textbf{.415} & \textbf{.458} & \textbf{.416} \\
\bottomrule
\end{tabular}
\label{tab:ablation}
\end{table}

\begin{table}[t]
\centering
\small
\caption{Masked Language Modeling Performance of Trait-Level Sub-Agents Across OCEAN Traits. 
}
\label{tab:mlm}
\setlength{\tabcolsep}{6pt}
\begin{tabular}{l c c c}
\toprule
\textbf{Agent} & \textbf{Test Acc.}$\uparrow$ & \textbf{Acc@5}$\uparrow$ & \textbf{Perplexity}$\downarrow$ \\
\midrule
O Baseline & 9.02\%  & 11.73\% & 8.51 \\
O High     & 44.68\% & 77.81\% & 6.81 \\
O Neutral  & 49.63\% & 78.52\% & 6.90 \\
O Low      & 61.32\% & 83.52\% & 6.84 \\
\midrule
C Baseline & 10.06\% & 13.82\% & 7.73 \\
C High     & 50.72\% & 76.71\% & 7.46 \\
C Neutral  & 63.41\% & 83.37\% & 7.16 \\
C Low      & 50.93\% & 80.95\% & 7.58 \\
\midrule
E Baseline & 9.07\%  & 11.84\% & 8.47 \\
E High     & 53.53\% & 78.45\% & 7.41 \\
E Neutral  & 48.92\% & 81.85\% & 6.83 \\
E Low      & 53.71\% & 78.65\% & 7.19 \\
\midrule
A Baseline & 9.41\%  & 12.52\% & 8.22 \\
A High     & 57.96\% & 82.42\% & 6.39 \\
A Neutral  & 52.94\% & 79.60\% & 7.50 \\
A Low      & 48.28\% & 78.12\% & 6.68 \\
\midrule
N Baseline & 8.95\%  & 11.59\% & 8.57 \\
N High     & 46.38\% & 79.84\% & 7.20 \\
N Neutral  & 56.07\% & 82.01\% & 6.84 \\
N Low      & 52.55\% & 78.64\% & 6.76 \\

\bottomrule
\end{tabular}
\end{table}

\subsection{MLM Fine-tuning Evaluation}
Table~\ref{tab:mlm} shows MLM performance of fine-tuned trait-level sub-agents compared to a pretrained baseline. The baseline reflects the average performance of predicting masked tokens across HIGH, LOW, and NEUTRAL tasks without any fine-tuning, achieving only 9–10\% Test Accuracy and 12–14\% Acc@5. Fine-tuned HIGH, LOW, and NEUTRAL sub-agents substantially outperform this baseline, achieving an average Test Accuracy of 52.7\% and Acc@5 of 80.2\%. Perplexity also decreases, indicating more confident predictions after MLM fine-tuning. 

HIGH and LOW agents consistently outperform NEUTRAL, reflecting the inherent difficulty of predicting tokens in moderate narratives. These results demonstrate that MLM fine-tuning effectively instills trait-specific language patterns, supporting downstream multi-agent personality inference.

\subsection{MLM Visualization}

We visualize the learned personality-specific adaptations of our sub-agents by projecting their fine-tuned LoRA weights into a 2D space using Principal Component Analysis (PCA). Each point in the plot represents the task vector $\tau$ (Eq.\ref{eq:1}), of a sub-agent computed after training on a specific personality trait level. It is the difference between its fine-tuned and base model weights. This approach mirrors prior work on visual LoRAs \citep{liu2024lora}, showing that LoRA parameters themselves form a faithful embedding of task-specific behavior, enabling systematic analysis of trained sub-agents.

\begin{figure}[t]
\centering
\includegraphics[width=0.45\textwidth]{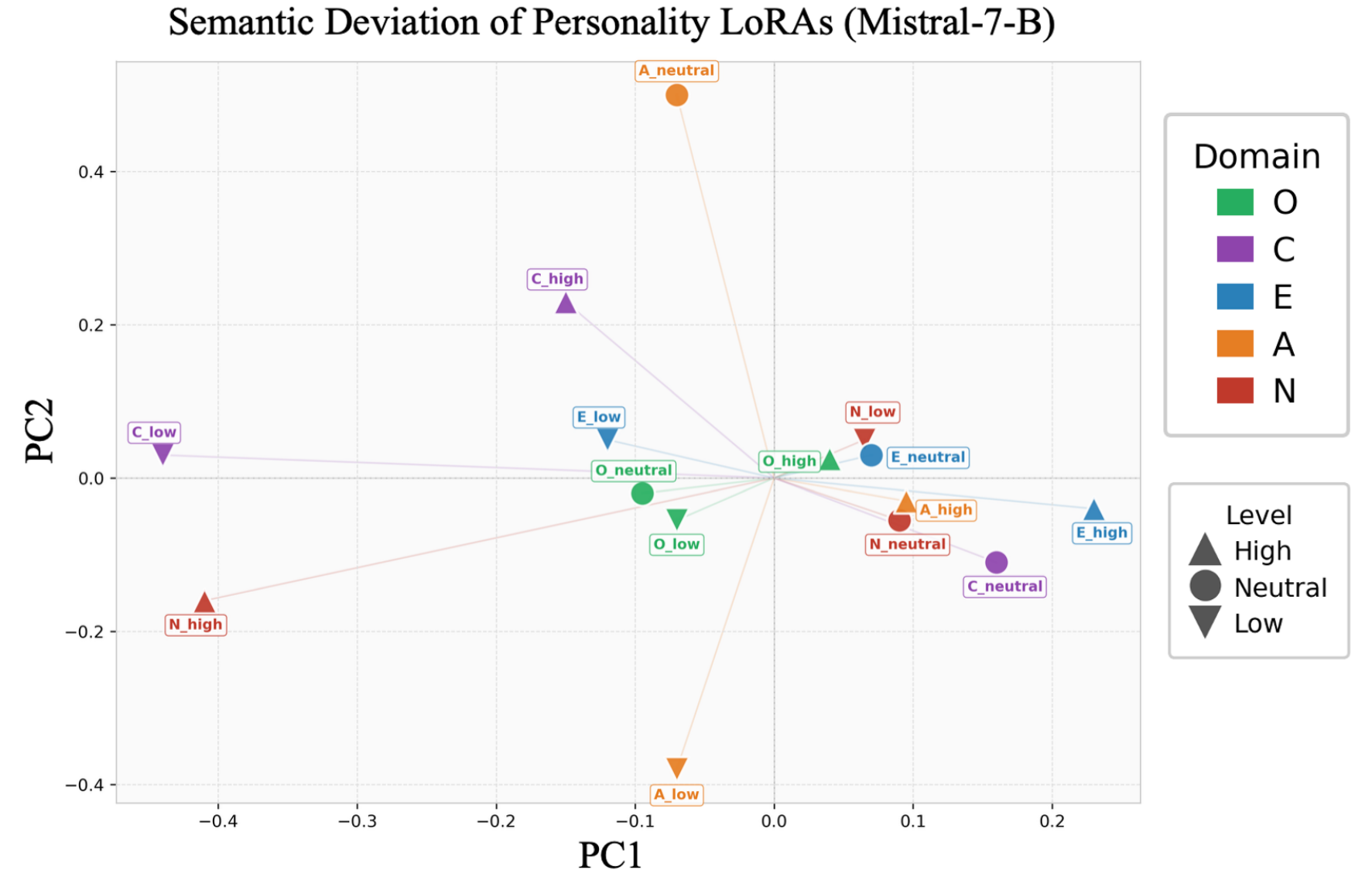}
\caption{Semantic separation of personality-specialized sub-agents after MLM fine-tuning.}
\label{fig:semantic_deviation}
\end{figure}

As shown in Figure~\ref{fig:semantic_deviation}, sub-agents trained on different traits are well-separated in PCA space. Even sub-agents representing different levels within the same trait are distinct, indicating that LoRA fine-tuning captures meaningful trait-specific information and that task vectors encode semantically relevant differences. Some proximity is observed between related trait levels. For example, low Neuroticism (N\_low) is near high Openness (O\_high) and neutral Extraversion (E\_neutral), while high agreeableness (A\_high) is closer to neutral Neuroticism (N\_neutral). These patterns align with broader tendencies in personality structure, where certain traits can co-vary or exhibit semantic overlap in behavioral expression \citep{soto2017bfi2}. Such semantic similarity across trait-level expressions can blur distinctions in embedding space. MLM fine-tuning provides a baseline for personality adaptation, and psychometric grounding through facet-level prompts further refines trait-level differentiation where needed.


\subsection{Qualitative Analysis}
\subsubsection{Semantic Alignment}  
\label{sec:ipip_alignment}






To quantify semantic alignment, we measure how closely sub-agent explanations correspond to the intended trait orientation by computing the cosine similarity between each sub-agent’s evidence $e_t^i$ and the High and Low IPIP-NEO facet keys for the corresponding trait. Embeddings are obtained using the Sentence Transformer all-mpnet-base-v2 \cite{reimers-2019-sentence-bert}.
 
For each trait $t$ and sub-agent $A_t^i \in \{\text{HIGH}, \text{LOW}\}$, $e_t^i$ denote the collected evidence (Algorithm \ref{alg:fado}). 
$\mathcal{F}_{t}^{H}$ and $\mathcal{F}_{t}^{L}$ represent the High and Low IPIP-NEO facet key sets for trait $t$, respectively. We compute cosine similarities as:
\[
s_H = \text{Sim}(e_t^i, \mathcal{F}_{t}^{H}), 
\qquad
s_L = \text{Sim}(e_t^i, \mathcal{F}_{t}^{L})
\]

The directional semantic alignment margin is defined as:
\[
\delta_t^i =
\begin{cases}
s_H - s_L, & \text{if } A_t^i \text{ is HIGH}, \\
s_L - s_H, & \text{if } A_t^i \text{ is LOW}.
\end{cases}
\]

Positive $\delta_t^i$ indicates correct semantic orientation toward the intended trait level. Neutral agents are excluded from margin computation since they use a blended combination of positive and negative keys. 

Table~\ref{tab:ipip_alignment} compares directional semantic alignment across the full model and its ablations to highlight how different components contribute to collecting trait-relevant evidence. The full multi-agent framework achieves the highest alignment margins across all traits, demonstrating effective personality induction. Removing MLM fine-tuning or trait-level role prompts moderately reduces alignment, suggesting that these components help sub-agents adopt trait-consistent behaviors but do not directly drive evidence collection aligned with IPIP-NEO keys. Omitting IPIP-NEO facet keys causes the largest drops, with some agents even showing negative margins, highlighting the critical role of explicit facet-level grounding in collecting trait-aligned evidence. Conscientiousness shows smaller sensitivity to facet-level cues, consistent with Table \ref{tab:baselines}, likely due to its highly cohesive facets that require less fine-grained decomposition.

\begin{table}[t]
\centering
\small
\caption{Directional semantic alignment between sub-agent explanations and IPIP-NEO facet keys across model configurations. Larger positive scores indicate stronger alignment with the intended trait direction.}
\label{tab:ipip_alignment}
\begin{tabular}{l c c c c}
\toprule
\textbf{Agent} & \textbf{Full} & \textbf{No} & \textbf{No Trait-} & \textbf{No IPIP-} \\
 & \textbf{Model} & \textbf{MLM} & \textbf{Level Role} & \textbf{Facet Keys} \\
\midrule
O\_high & +0.023 & +0.003& +0.005& -0.005\\
O\_low  & +0.089 & +0.012& +0.015& +0.009\\
\midrule
C\_high & +0.032 & +0.015& +0.030& +0.033\\
C\_low  & +0.026 & +0.014& -0.017& -0.002\\
\midrule
E\_high & +0.041 & +0.027& -0.001& -0.003\\
E\_low  & +0.026 & +0.018& +0.014& +0.058\\
\midrule
A\_high & +0.106 & +0.091& +0.047& +0.007\\
A\_low  & +0.017 & +0.050& +0.031& -0.026\\
\midrule
N\_high & +0.012 & +0.006& -0.025& -0.060\\
N\_low  & +0.134 & +0.014& +0.112& 0.015\\
\bottomrule
\end{tabular}
\end{table}

\begin{figure}[!t]
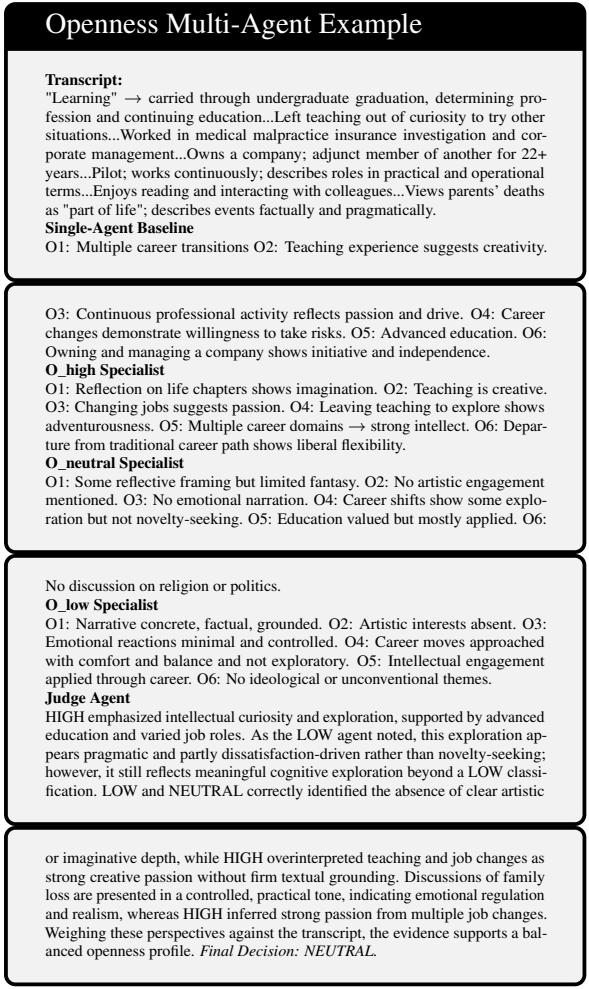

\centering
\begin{tcolorbox}[colback=gray!10,colframe=black,
title=Openness Multi-Agent Example,
breakable, width=\linewidth]
\tiny

\textbf{Transcript:}

"Learning" → carried through undergraduate graduation, determining profession and continuing education...Left teaching out of curiosity to try other situations...Worked in medical malpractice insurance investigation and corporate management...Owns a company; adjunct member of another for 22+ years...Pilot; works continuously; describes roles in practical and operational terms...Enjoys reading and interacting with colleagues...Views parents' deaths as "part of life"; describes events factually and pragmatically.

\textbf{Single-Agent Baseline}

O1: Multiple career transitions 
O2: Teaching experience suggests creativity. 
O3: Continuous professional activity reflects passion and drive.
O4: Career changes demonstrate willingness to take risks.
O5: Advanced education.
O6: Owning and managing a company shows initiative and independence.

\textbf{O\_high Specialist}

O1: Reflection on life chapters shows imagination.  
O2: Teaching is creative. 
O3: Changing jobs suggests passion.
O4: Leaving teaching to explore shows adventurousness.  
O5: Multiple career domains → strong intellect.  
O6: Departure from traditional career path shows liberal flexibility.

\textbf{O\_neutral Specialist}

O1: Some reflective framing but limited fantasy.  
O2: No artistic engagement mentioned.  
O3: No emotional narration.  
O4: Career shifts show some exploration but not novelty-seeking.  
O5: Education valued but mostly applied.  
O6: No discussion on religion or politics.

\textbf{O\_low Specialist}

O1: Narrative concrete, factual, grounded.  
O2: Artistic interests absent.  
O3: Emotional reactions minimal and controlled.  
O4: Career moves approached with comfort and balance and not exploratory.  
O5: Intellectual engagement applied through career.  
O6: No ideological or unconventional themes.

\textbf{Judge Agent} 

HIGH emphasized intellectual curiosity and exploration, supported by advanced education and varied job roles. As the LOW agent noted, this exploration appears pragmatic and partly dissatisfaction-driven rather than novelty-seeking; however, it still reflects meaningful cognitive exploration beyond a LOW classification. LOW and NEUTRAL correctly identified the absence of clear artistic or imaginative depth, while HIGH overinterpreted teaching and job changes as strong creative passion without firm textual grounding. Discussions of family loss are presented in a controlled, practical tone, indicating emotional regulation and realism, whereas HIGH inferred strong passion from multiple job changes. Weighing these perspectives against the transcript, the evidence supports a balanced openness profile.
\emph{Final Decision: NEUTRAL.}

\end{tcolorbox}
\caption{Multi-agent example for Openness.}
\label{fig:openness_conflict_transcript}
\end{figure}


\subsubsection{Use-Case}
To further illustrate the benefits of the proposed multi-agent framework, we present a use-case in Figure~\ref{fig:openness_conflict_transcript} where a single-agent model fails while the judge-mediated multi-agent system succeeds. In this Openness case study, the HIGH sub-agent emphasizes intellectual curiosity, career exploration, and perceived creativity, while the LOW and NEUTRAL sub-agents highlight the pragmatic, factual tone of the narrative and the absence of artistic or imaginative depth aligned with their specific trait-level. The NEUTRAL sub-agent also notes some exploration but points out limited novelty-seeking. The judge agent integrates these contrasting viewpoints, and weighs textual evidence from the full transcript to correctly assign a NEUTRAL classification. By contrast, a single-agent baseline classifies this profile as HIGH openness, relying primarily on surface cues such as multiple career transitions, advanced education, and self-reported curiosity, without distinguishing between pragmatic or adventure-seeking career mobility. This example demonstrates that multi-agent reasoning effectively reconciles divergent inferences from specialized sub-agents, mitigating over- or under-interpretation and producing robust, contextually grounded personality assessments.

\section{Discussion}


Our experiments illustrate how a structured multi-agent framework transforms personality assessment into a more robust and reliable reasoning process. By decomposing inference across HIGH, LOW, and NEUTRAL sub-agents, the system separates trait-level specific signals from contextual confounds, enabling the final judgment to distinguish between surface-level cues and deeper personality indicators. For instance, in the example in Figure \ref{fig:openness_conflict_transcript}, exploratory career transitions may appear as strong markers of Openness to a single-agent model, but when examined alongside LOW and NEUTRAL sub-agents, the framework recognizes pragmatic motivations and moderates the final judgment. This demonstrates that multi-agent reasoning is not merely an ensemble of predictions, but it strategically reconciles different interpretations to produce more accurate outputs.

The ablation and semantic alignment results further clarify why this structure matters. Prior work has shown that role prompting can induce persona-consistent behavior in LLMs \cite{yeo2024pado}, but such induction can be limited. Our ablation analysis shows that each component of our model contributes to overall performance and personality induction. Our alignment analysis shows that consistent polarity in HIGH and LOW agents depends critically on facet-level IPIP-NEO facet keys in the prompt. When these are removed, directional alignment weakens substantially. This indicates that Role prompts and fine-tuning shape expression strength, while psychometric supervision guides sub-agents to avoid generic interpretations.


Overall, the quantitative results and qualitative analyses demonstrate that reliable personality inference from long, complex text with subtle expressions requires both decomposition and psychometric grounding. This approach enables robust modeling of nuanced, context-dependent psychological signals in text, underscoring the value of integrating structured psychological theory with LLM-based multi-agent reasoning.

\section{Conclusion}
We present a multi-agent framework for text-based personality assessment that combines trait-specific sub-agents, role prompts, and IPIP-NEO facet prompts. Sub-agent specialization provides competing perspectives, psychometric alignment maintains theoretical coherence, and a judge integrates them into predictions that are more accurate than single-agent baselines. Results demonstrate robust performance, improved handling of conflicting cues, and clear trait-aligned explanations.


\section{Limitations}
While the framework shows strong performance, several limitations remain. First, moderate NEUTRAL expressions are harder to classify, reflecting the inherent ambiguity of intermediate trait manifestations in narratives. Second, the approach relies on transcript data collected in a clinical setting, which may limit generalization to noisier or less structured text. Third, although our multi-agent setup improves reliability, it increases computational cost compared to single-agent models. Finally, the current evaluation is limited to English-language IPIP-NEO based narratives; cross-lingual or culturally distinct adaptations remain unexplored.

\section{Ethics Statement}
Our work focuses on automated personality assessment for research purposes and emphasizes transparency, interpretability, and psychometric grounding. All data used are anonymized and derived from consented participants. We caution that any automated personality assessment carries risks of misinterpretation or bias if deployed in sensitive contexts such as hiring or mental health evaluation. The framework is intended to support understanding of narrative patterns and cognitive traits, not to make high-stakes decisions without human oversight. Researchers and practitioners should apply these models responsibly and consider ethical implications in deployment.

\bibliography{fado_refs}

\appendix

\section{IPIP-NEO Facet Keys}
\label{app:facets}
Table \ref{tab:facets_all} shows complete list of IPIP-NEO facet positive and negative keys used for HIGH and LOW agents, respectively. For NEUTRAL agents, we created moderate cues by blending the semantic content of positive and negative keys, keeping core behaviors but removing extreme qualifiers or reducing intensity. This produces balanced, mid-level expressions that reflect natural trait variation without compromising psychometric validity.

\begin{table*}[t]
\caption{IPIP-NEO facet keys for HIGH, LOW, and NEUTRAL sub-agents. NEUTRAL agents use blended descriptions reflecting moderate expression.}
\label{tab:facets_all}
\centering
\tiny
\begin{tabular}{p{2.2cm}p{4cm}p{4cm}p{4cm}}
\toprule
\textbf{Facet} & \textbf{HIGH} & \textbf{LOW} & \textbf{NEUTRAL} \\
\midrule
\multicolumn{4}{l}{\textit{Neuroticism}} \\
N1 Anxiety          & Worry about things, fear for the worst, get stressed out easily
                    & Relaxed most of the time, not easily disturbed
                    & Occasionally worry about things, generally handle stress well \\
N2 Anger            & Get angry easily, often in a bad mood, lose my temper
                    & Rarely get irritated, keep my cool
                    & Sometimes get frustrated, usually maintain composure \\
N3 Depression       & Often feel blue, dislike myself, low opinion of myself
                    & Seldom feel blue, feel comfortable with myself
                    & Sometimes feel down, but generally feel okay about myself \\
N4 Self-Consciousness & Easily intimidated, difficult to approach others
                    & Not embarrassed easily, comfortable in new situations
                    & Occasionally self-conscious, can adapt in social situations \\
N5 Immoderation     & Do things I later regret, often overindulge
                    & Rarely overindulge, easily resist temptations
                    & Occasionally act impulsively, generally maintain self-control \\
N6 Vulnerability    & Panic easily, overwhelmed by events
                    & Remain calm under pressure, readily overcome setbacks
                    & Sometimes feel overwhelmed, but can recover with effort \\
\midrule
\multicolumn{4}{l}{\textit{Extraversion}} \\
E1 Friendliness     & Make friends easily, warm up quickly to others
                    & Hard to get to know, keep others at a distance
                    & Generally friendly, take some time to warm up \\
E2 Gregariousness   & Love large parties, enjoy being part of a group
                    & Prefer to be alone, avoid crowds
                    & Occasionally enjoy group settings, sometimes prefer solitude \\
E3 Assertiveness    & Take charge, try to lead others
                    & Keep in the background, hold back opinions
                    & Sometimes take initiative, other times stay reserved \\
E4 Activity         & Always busy, always on the go
                    & Like to take it easy, react slowly
                    & Occasionally busy, generally maintain moderate pace \\
E5 Excitement       & Love excitement, seek adventure
                    & Avoid danger, dislike loud and hectic settings
                    & Sometimes seek new experiences, usually avoid risky situations \\
E6 Cheerfulness     & Radiate joy, look at the bright side
                    & Not easily amused, seldom joke
                    & Usually cheerful, sometimes feel down or serious \\
\midrule
\multicolumn{4}{l}{\textit{Openness}} \\
O1 Imagination      & Vivid imagination, love to daydream
                    & Difficulty imagining things, seldom daydream
                    & Occasionally imagine or daydream, generally practical \\
O2 Artistic Interests & Love art, music, and challenging material
                    & Do not enjoy art, poetry, or music
                    & Sometimes engage with art or music, not consistently \\
O3 Emotionality     & Experience emotions intensely, feel others' emotions
                    & Not easily affected by emotions
                    & Occasionally affected by emotions, generally balanced \\
O4 Adventurousness  & Prefer variety to routine, try new foods
                    & Prefer familiar to new experiences
                    & Sometimes try new experiences, usually follow routine \\
O5 Intellect        & Love to read, enjoy complex thinking
                    & Avoid difficult reading and philosophy
                    & Sometimes enjoy reading or complex thinking, moderately curious \\
O6 Liberalism       & Flexible views, question absolutes
                    & Prefer conventional social and political views
                    & Open to new ideas occasionally, but hold some traditional views \\
\midrule
\multicolumn{4}{l}{\textit{Agreeableness}} \\
A1 Trust            & Believe others have good intentions
                    & Suspect hidden motives, distrust people
                    & Generally trust others, sometimes cautious \\
A2 Morality         & Do not take advantage of others
                    & Use flattery to get ahead, misrepresent facts
                    & Usually honest, occasionally pragmatic \\
A3 Altruism         & Love to help others, make people feel welcome
                    & Take little time for others
                    & Sometimes help others, usually neutral \\
A4 Cooperation      & Avoid imposing will on others
                    & Enjoy confrontation, contradict others
                    & Usually cooperative, occasionally assertive \\
A5 Modesty          & Do not believe I am better than others
                    & Think highly of myself, always know the answers
                    & Moderate self-confidence, sometimes assertive \\
A6 Sympathy         & Care for the feelings of others
                    & Indifferent to the feelings of others
                    & Sometimes attentive to others’ feelings, occasionally indifferent \\
\midrule
\multicolumn{4}{l}{\textit{Conscientiousness}} \\
C1 Self-Efficacy    & Complete tasks successfully, handle tasks smoothly
                    & Misjudge situations, little to contribute
                    & Usually able to complete tasks, sometimes uncertain \\
C2 Orderliness      & Like order, want everything just right
                    & Leave a mess, not bothered by disorder
                    & Generally organized, occasionally messy \\
C3 Dutifulness      & Follow rules, keep promises, tell the truth
                    & Break rules, avoid duties
                    & Generally dutiful, sometimes flexible with rules \\
C4 Achievement Striving & Work hard, do more than expected
                    & Do just enough to get by
                    & Sometimes strive for more, generally moderate effort \\
C5 Self-Discipline  & Get chores done right away, stay prepared
                    & Waste time, postpone decisions
                    & Usually disciplined, occasionally procrastinate \\
C6 Caution          & Avoid mistakes, choose words with care
                    & Jump into things without thinking
                    & Sometimes cautious, sometimes act spontaneously \\
\bottomrule
\end{tabular}
\end{table*}

\section{Prompt Structure}\label{app:prompt}
Figure \ref{fig:subagent_prompt} and \ref{fig:judge_prompt} show templates for sub-agent and judge prompts, respectively.

\begin{figure*}[t]
\centering
\tiny  
\begin{tcolorbox}[colback=gray!10,colframe=black,title=Sub-agent Prompt, 
                  width=\textwidth, enlarge left by=0mm, enlarge right by=0mm]

\textbf{Role Prompt:}

You are a domain expert in the assessment of \{\{TRAIT LEVEL\}\} \{\{TRAIT NAME\}\} within the framework of the IPIP-NEO personality inventory. 

\{\{TRAIT LEVEL DESCRIPTION\}\} 

Your role is to systematically analyze interview transcripts to identify behavioral patterns, verbal cues, and contextual indicators that reflect \{\{TRAIT NAME\}\}. The focus should be on evidence-based evaluation aligned with established IPIP-NEO facet key definitions.

\textbf{Analysis Task:}

Examine the following interview transcript to evaluate \{\{TRAIT NAME\}\} at the \{\{TRAIT LEVEL\}\} level. For each of the six IPIP-NEO facets, you are provided with facet-specific keys that correspond to the assigned trait level:  

\begin{enumerate}[label=\arabic*.]
    \item \{\{FACET 1 NAME\}\}: \{\{FACET 1 Keys\}\}
    \item \{\{FACET 2 NAME\}\}: \{\{FACET 2 Keys\}\}
    \item \{\{FACET 3 NAME\}\}: \{\{FACET 3 Keys\}\}
    \item \{\{FACET 4 NAME\}\}: \{\{FACET 4 Keys\}\}
    \item \{\{FACET 5 NAME\}\}: \{\{FACET 5 Keys\}\}
    \item \{\{FACET 6 NAME\}\}: \{\{FACET 6 Keys\}\}
\end{enumerate}

\textbf{Instructions:}
\begin{enumerate}[label=\arabic*.]
    \item For each facet, identify evidence in the transcript that aligns with the provided keys, which are specifically selected to reflect the \{\{TRAIT LEVEL\}\} level.
    \item Synthesize the facet-level evidence to make an overall evaluation: Does the individual demonstrate \{\{TRAIT LEVEL (UPPERCASE)\}\} \{\{TRAIT NAME\}\}? Support your conclusion with explicit references to the transcript.
\end{enumerate}

\textbf{Interview Transcript:} \{\{TRANSCRIPT\}\}

\textbf{Output:}

Generate ONLY valid JSON in the exact format below. Do not include any explanations or preamble.

\begin{verbatim}
{
  "facet_analysis": {
    "{{FACET 1 KEY}}": "evidence from the transcript supporting this facet keys",
    "{{FACET 2 KEY}}": "evidence from the transcript supporting this facet keys",
    "{{FACET 3 KEY}}": "evidence from the transcript supporting this facet keys",
    "{{FACET 4 KEY}}": "evidence from the transcript supporting this facet keys",
    "{{FACET 5 KEY}}": "evidence from the transcript supporting this facet keys",
    "{{FACET 6 KEY}}": "evidence from the transcript supporting this facet keys"
  },
  "overall_evaluation": {
    "determination": "YES or NO",
    "confidence": "High / Medium / Low",
    "reasoning": "Concise synthesis of facet-level evidence (2–3 sentences)"
  }
}
\end{verbatim}

\textbf{Note:} Output JSON only.

\end{tcolorbox}
\caption{Sub-agent prompt template.}
\label{fig:subagent_prompt}
\end{figure*}





\begin{figure*}[t]
\centering
\scriptsize  
\begin{tcolorbox}[colback=gray!10,colframe=black,title=JUDGE AGENT — FINAL TRAIT SYNTHESIS (IPIP-NEO / Big Five), 
                  width=\textwidth, enlarge left by=0mm, enlarge right by=0mm]
\textbf{Role:}

You are a domain expert in personality assessment using the Big Five (OCEAN) model within the IPIP-NEO framework. Your objective is to review 15 expert sub-agent 3 per trait outputs derived from the same interview transcript and synthesize them into final trait determinations for: Neuroticism, Extraversion, Openness, Agreeableness, and Conscientiousness.

\textbf{Inputs:}
\begin{itemize}
    \item Original interview transcript.
    \item Fifteen expert sub-agent outputs, consisting of HIGH, LOW, and NEUTRAL assessments for each trait.
\end{itemize}

\textbf{Trait Composition (IPIP-NEO Facets):}
\begin{itemize}
    \item \textbf{Neuroticism}: Anxiety, Anger, Depression, Self-Consciousness, Immoderation, Vulnerability
    \item \textbf{Extraversion}: Friendliness, Gregariousness, Assertiveness, Activity Level, Excitement Seeking, Cheerfulness
    \item \textbf{Openness}: Imagination, Artistic Interests, Emotionality, Adventurousness, Intellect, Liberalism
    \item \textbf{Agreeableness}: Trust, Morality, Altruism, Cooperation, Modesty, Sympathy
    \item \textbf{Conscientiousness}: Self-Efficacy, Orderliness, Dutifulness, Achievement Striving, Self-Discipline, Cautiousness
\end{itemize}

\textbf{Analysis Task:}
\begin{enumerate}[label=\arabic*.]
    \item Review all three sub-agent outputs (HIGH, LOW, NEUTRAL) for each trait.
    \item Evaluate whether the facet-level evidence aligns with the canonical structure of the trait.
    \item Resolve conflicts by considering facet coverage, specificity of transcript evidence, and internal consistency of reasoning.
    \item Integrate the findings to produce a single final determination for each trait: HIGH, LOW, or NEUTRAL.
\end{enumerate}

\textbf{Output Format:}

Generate ONLY valid JSON in the exact format below. Each entry should synthesize facet-level evidence from the sub-agent outputs and reflect your final judgment. Do not include explanations or commentary outside the JSON.

\begin{verbatim}
{
  "neuroticism": {
    "final_determination": "HIGH / LOW / NEUTRAL",
    "confidence": "High / Medium / Low",
    "reasoning": "Facet-level synthesis explaining why the evidence
    best supports this level (2–3 sentences)",
    "votes": Summary: N_high=YES/NO, N_low=YES/NO, N_neutral=YES/NO
  },
  "extraversion": {
    "final_determination": "HIGH / LOW / NEUTRAL",
    "confidence": "High / Medium / Low",
    "reasoning": "Facet-level synthesis explaining why the evidence
    best supports this level (2–3 sentences)",
    "votes": "Summary: E_high=YES/NO, E_low=YES/NO, E_neutral=YES/NO"
  },
  "openness": {
    "final_determination": "HIGH / LOW / NEUTRAL",
    "confidence": "High / Medium / Low",
    "reasoning": "Facet-level synthesis explaining why the evidence
    best supports this level (2–3 sentences)",
    "votes": "Summary: O_high=YES/NO, O_low=YES/NO, O_neutral=YES/NO"
  },
  "agreeableness": {
    "final_determination": "HIGH / LOW / NEUTRAL",
    "confidence": "High / Medium / Low",
    "reasoning": "Facet-level synthesis explaining why the evidence
    best supports this level (2–3 sentences)",
    "votes": "Summary: A_high=YES/NO, A_low=YES/NO, A_neutral=YES/NO"
  },
  "conscientiousness": {
    "final_determination": "HIGH / LOW / NEUTRAL",
    "confidence": "High / Medium / Low",
    "reasoning": "Facet-level synthesis explaining why the evidence
    best supports this level (2–3 sentences)",
    "votes": "Summary: C_high=YES/NO, C_low=YES/NO, C_neutral=YES/NO"
  }
}
\end{verbatim}

\textbf{Note:} Output JSON only. Do not include any preamble or explanation.
\end{tcolorbox}
\caption{Judge agent prompt template.}
\label{fig:judge_prompt}
\end{figure*}

\section{Implementation Details}\label{apx:imp}

All fifteen sub-agents are initialized from \texttt{Mistral-7B-Instruct}, while the judge model uses \texttt{Qwen}, chosen for its larger context window necessary to aggregate evidence from multiple agents.

Each sub-agent is fine-tuned using masked language modeling (MLM), with LoRA applied to the \texttt{q\_proj}, \texttt{k\_proj}, \texttt{v\_proj}, and \texttt{o\_proj} layers. LoRA hyperparameters are $r=4$, $\alpha=8$, and dropout 0.05, resulting in approximately 4.2M trainable parameters per agent. Training uses AdamW (learning rate $2\times10^{-4}$), batch size 64 with gradient accumulation, for 500–800 steps on a single NVIDIA A100 GPU.

We define the key parameter sets as follows: $\theta_\mathrm{base}$ for the pre-trained sub-agent parameters, $\theta_\mathrm{ft}$ for the fine-tuned parameters after LoRA adaptation, and the task vector $\tau = \theta_\mathrm{ft} - \theta_\mathrm{base}$. Task vectors are merged back into the base model using
\begin{equation}
\theta_\mathrm{merged} = \theta_\mathrm{base} + \lambda \, \tau, \quad \lambda = 0.3
\end{equation}
We found that larger $\lambda$ degraded structured JSON generation, while smaller values reduced trait-specific adaptation.

At inference, sub-agents run sequentially in fp16, clearing GPU memory between runs, while the judge model remains loaded to aggregate predictions.

LoRA Fine-Tuning Hyperparameters are shown in table \ref{tab:lora_config}.

\begin{table}[h]
\centering
\small
\caption{LoRA hyperparameters for all fifteen sub-agents.}
\begin{tabular}{lc}
\toprule
\textbf{Parameter} & \textbf{Value} \\
\midrule
Fine-tuning base model       & Mistral-7B-Instruct-v0.2 \\
Baseline sub-agent model     & Mistral-7B-Instruct-v0.3 \\
Judge model                  & Qwen2.5-7B-Instruct \\
\midrule
LoRA rank ($r$)              & 4 \\
LoRA $\alpha_{\text{LoRA}}$  & 8 \\
LoRA dropout                 & 0.05 \\
Target modules               & q\_proj, k\_proj, v\_proj, o\_proj \\
Trainable parameters         & approximately 4.2M \\
Merge coefficient ($\alpha$) & 0.30 \\
Training steps               & 500 to 800 \\
Global batch size            & 64 \\
Learning rate                & $2 \times 10^{-4}$ \\
Optimizer                    & AdamW \\
Window stride                & 50 percent overlap \\
Precision                    & float16 \\
GPU                          & A100-SXM4-80GB \\
Time per model               & approximately 8 to 10 hours \\
\bottomrule
\end{tabular}

\label{tab:lora_config}
\end{table}

\end{document}